\documentclass{article}

\usepackage{style}
\usepackage{courier}
\usepackage{libertine}
\usepackage[utf8]{inputenc} 
\usepackage[T1]{fontenc}    
\usepackage{hyperref}       
\usepackage{url}            
\usepackage{booktabs}       
\usepackage{amsfonts}       
\usepackage{nicefrac}       
\usepackage{microtype}      
\usepackage{xcolor}         

\usepackage{graphicx}
\usepackage{amsmath}
\usepackage{amssymb}
\usepackage{listings}
\usepackage{algorithm}
\usepackage{algpseudocode}%
\usepackage{wrapfig, blindtext}

\newcommand{\code}[1]{\texttt{\small{#1}}}

\usepackage[square,numbers]{natbib}
\usepackage{sourcecodepro}

\title{Adaptable Agent Populations\\via a Generative Model of Policies}

\author{%
  Kenneth Derek\\
  MIT CSAIL\\
  \small{\texttt{kderek@alum.mit.edu}} \\
  \and
   Phillip Isola \\
   MIT CSAIL\\
   \small{\texttt{phillipi@mit.edu}} \\
}

\begin{document}

\date{\vspace{-3ex}}

\maketitle

\begin{abstract}
In the natural world, life has found innumerable ways to survive and often thrive. Between and even within species, each individual is in some manner unique, and this diversity lends adaptability and robustness to life. In this work, we aim to learn a space of diverse and high-reward policies on any given environment. To this end, we introduce a generative model of policies, which maps a low-dimensional latent space to an agent policy space. Our method enables learning an entire population of agent policies, without requiring the use of separate policy parameters. Just as real world populations can adapt and evolve via natural selection, our method is able to adapt to changes in our environment solely by selecting for policies in latent space. We test our generative model’s capabilities in a variety of environments, including an open-ended grid-world and a two-player soccer environment. Code, visualizations, and additional experiments can be found at \texttt{\href{https://kennyderek.github.io/adap/}{https://kennyderek.github.io/adap/}}.
\end{abstract}

\section{Introduction}

Quick thought experiment: imagine our world was such that all people acted, thought, and looked \textit{exactly} the same in \textit{every} situation. Would we ever have found the influential dissenters that sparked scientific, political, and cultural revolutions?

In reinforcement learning (RL), it is common to learn a single policy that fits an environment. However, it is often desirable to instead find an entire array of high performing policies. To this end, we propose learning a generative model of policies. At a high level, we aim to show that purposefully learning a diverse policy space for a given environment can be competitive to learning a single policy, while better encompassing a range of skillful behaviors that are adaptable and robust to changes in the task and environment. We name our method of learning a space of \textit{ad}aptable \textit{a}gent \textit{p}olices: ADAP.

Previous work has touched on ideas akin to a generative model of policies.  In hierarchical RL, the the high-level policy controller can be considered a generator of sub-policies that are `options' \cite{sutton, diayn, florensa17}. But these methods are designed with the intent to find decomposable skills that aid in the construction of just one downstream controller policy. A body of prior work that aligns more closely with our goals is that of quality diversity \cite{qualitydiv}, which optimizes a population of agents along the axes of both reward and diversity. These methods often use evolutionary search and a require a discrete-sized population of separate agents, each with their own policy weights. This consumes more time and training resources, and limits the number of potential behaviors. Our work integrates the goals of quality diversity into time and memory efficient deep RL by simulating an entire population of agents via a generative model of policies, with diversity bounded only by capacity of the generator.

Why should we bother with finding more than one policy per environment? We propose two primary reasons. First, RL environments are continually approaching greater levels of open-endedness and complexity. For a given environment, there might be an entire manifold of valid and near-equally high performing strategies. By finding points across this manifold, we avoid `having all eggs in one basket,' granting robustness and adaptability to environmental changes. In the event of a change, we are able to adapt our generated population to select individuals that can still survive given the ablation, much like natural selection drives evolution in the real world. Secondly, using a generative model of policies as a population of agents makes intuitive sense in multi-agent environments, in which different agents should have the capacity to act like they are unique individuals. However, it is common in many multi-agent reinforcement learning settings to deploy the same policy across all agents, such that they are essentially distributed clones. Doing so may reduce the multi-modality of the agent population, resulting a single `average' agent.

The rest of the paper is organized as follows. First we introduce our generative model of policies and the diversity objective that guides its learning. Next, we explore the potentials of learning a population of agents by ablating environments and then searching for suitable policies, directly in latent space. We primarily study two environments: Markov Soccer \cite{markovsoccer} and Farmworld. Farmworld is a new environment we have developed for testing diversity in a multi-agent, open-ended gridworld. At the website linked in the abstract, one can find qualitative results of experiments presented in this paper, as well as additional results on toy environments of CartPole \cite{openaigym} and a standard multi-goal environment.

\section{Method} \label{2}

We learn a mapping $G: \phi, Z \rightarrow \Pi$, from generator weights ${\phi}$ and latent distribution $Z$ to a space of policies $\Pi$. The generator $G_{\phi}$ itself is not a policy, and must be conditioned on a $z \sim Z$ in order to define a learned set of behaviors that map states to actions. The latent vector $z$ is a stochastic parameter of $G_{\phi}$, and is sampled once at the beginning of each agent episode. In our experiments, we sample $z$ from the three dimensional unit sphere, which is low dimensional enough to quickly search the latent space. Note that unlike \cite{hypernets}, we do not need to build the weights of each agent policy, allowing memory efficiency when training many agents simultaneously. In practice, to compute agent actions, we provide the current environment observation and the agent's latent $z$ as inputs to $G_{\phi}$. The weights $\phi$ are differentiable with respect to our diversity regularizer \eqref{eq:divobjkl} and agent policy actions, and we update $\phi$ like normal policy weights. Generally, we drop the $\phi$ from notation for simplicity.

\begin{figure}[h]
\includegraphics[width=0.8\linewidth]{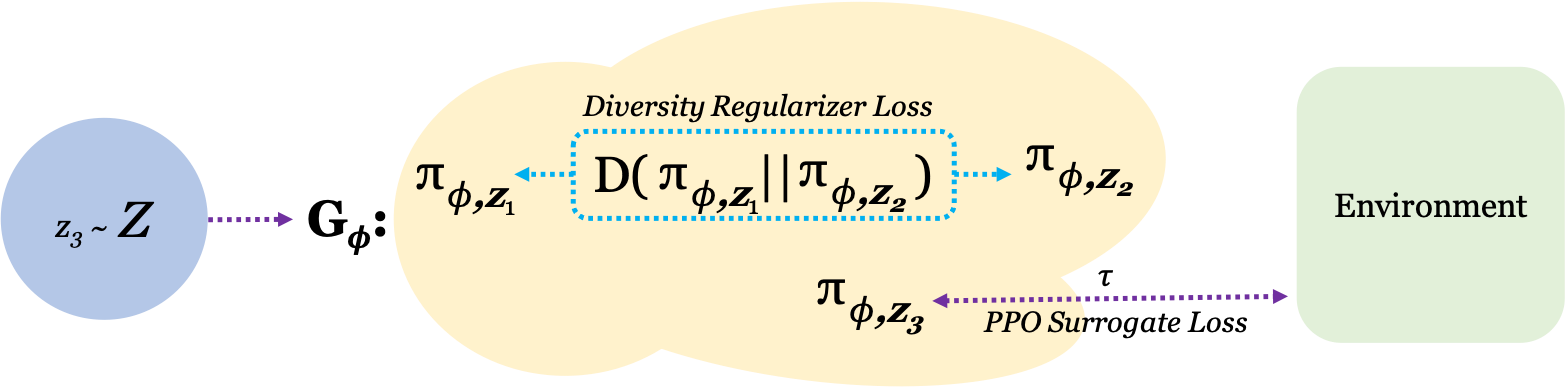}
\centering
\caption[Method Diagram]%
{Method Diagram. Upon agent initialization, we sample a latent $z \in Z$ which, along with $\phi$, defines an agent policy for the episode. The diversity regularizer loss is defined over the expected pair $(z_i, z_j)$ and is thus optimized over all possible policies -- not just the current policy used on the environment.}
\label{method}
\vspace{-0.2cm}
\end{figure}

\paragraph{Diversity Regularization}

In order to learn a diverse space of unique policies, we introduce a diversity regularization objective. Since policies define a space of actions taken over different states, we propose that in order for two policies to be distinct, they must have different action distributions given the same state. To this end, we define the objective $L_{div}$ \eqref{eq:divobjkl}:
\begin{equation}\label{eq:divobjkl}
    L_{div}(\phi) = 
    \mathop{\mathbb{E}}_{s \in S}
    \left[
        \mathop{\mathbb{E}}_{\substack{z_i, z_j \in Z \\ z_i \neq z_j}}
            \exp \left( -D_{KL}(\pi_{\phi,z_i;b}(s) \Vert \pi_{\phi,z_j;b}(s)) \right)
    \right]
\end{equation}
in which $D_{KL}$ is the KL-divergence between the two policy action distributions $\pi_{\theta, z_i}$ and $\pi_{\theta, z_j}$, and $b$ is a smoothing constant over the action distributions. We use the smoothing constant $b$ because otherwise, the KL-divergence may quickly approach infinity as probability mass over a particular action approaches zero (see the Appendix for more details).

\paragraph{Optimization of $G$}
In our experiments, we optimize the diversity objective in an online fashion using gradient descent, in conjunction with a PPO \cite{ppo} clipped-surrogate objective and an entropy regularization objective. Our full optimization problem is

$$\max_{\phi} L_{PPO}(\phi) - \alpha L_{div}(\phi)$$

where $L_{PPO}$ is Equation 9 in \cite{ppo} and $\alpha$ is a coefficient to scale the diversity regularization objective. See Algorithm \ref{ppo_div} for additional details.

We wrote and trained our models and algorithms using the RLLib framework \cite{rllib} distributed across 3 CPU workers. We ran at least three seeds for each experiment, of which the longest single trial ran for 30 million environment steps on CPU, training in under 24 hours using a single policy update worker and two rollout workers.

\paragraph{Adaptation via Optimization in the Latent Space of $G$} By learning an entire space of policies $\Pi$, we are able to search our policy space for the highest performing policy, whether dealing with the train environment or an ablated future environment.

In contrast to searching over policy parameters through transfer learning or fine-tuning, we are able to quickly search over the low-dimensional latent space (dimensionality 3 in our experiments). In fact, we can quickly adapt back and forth to various situations: the search procedure often takes \textit{less than 30 seconds}, or 100 episode rollouts, to find any high quality solutions that exist. Over the course of a small number of generations, we evaluate randomly sampled latents, and keep higher performing ones with greater probability. In the event that episodes have a high degree of variablility per run -- such as in the Markov Soccer environment -- it may be necessary to run several episodes per latent vector and average the returns. Details can be found in Algorithm \ref{lso}.

\begin{figure}[h]
\includegraphics[width=0.8\linewidth]{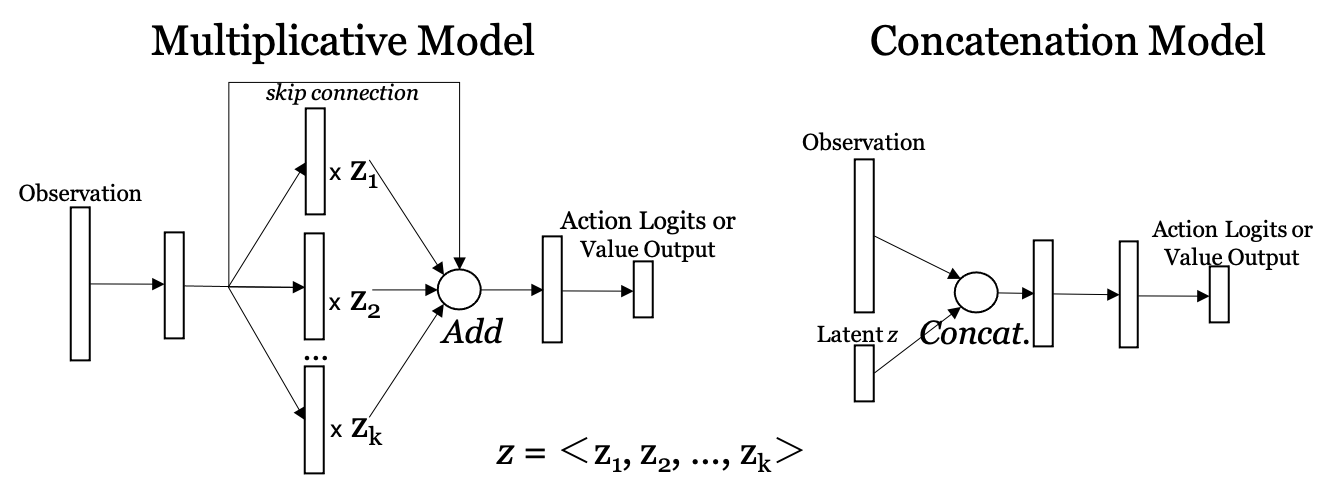}
\centering
\vspace{-0.2cm}
\caption{Model Architectures for Latent Integration}
\label{models}
\vspace{-0.4cm}
\end{figure}

\paragraph{Model Architecture} Similarly to prior work \cite{florensa17}, we have found that richer integrations between the latent vector and the observation can yield a more multi-modal policy space. To induce this richer integration, we introduce a multiplicative model for latent integration. Using a latent vector of dimension $k$, our multiplicative model is able to learn $k$ interpretations of the observation, which are each modulated by a dimension of the latent vector. A skip connection allows the model to learn policies faster than without. As a baseline, we use a concatenation model, in which the latent vector \textit{z} is concatenated with the environment observation at each timestep. In both cases, by setting corresponding model weights to zero, a learned policy could completely ignore the latent vector to yield a standard RL policy architecture.

Note that the multiplicative model architecture comes with increased computational cost, in which for a hidden dimension of size $d$ and latent dimension $k$, the number of parameters of the hidden layers are bounded by $\Theta((k+1)d^2)$, whereas in the concatenation model, they are bounded by $\Theta(d^2)$. In practice, since $k$ and $d$ are small ($k=3$ and $d\in \{16, 32, 64\}$) in our experiments, the increase in computational cost is not significant.

\section{Related Work}

\paragraph{Quality Diversity} The evolutionary computing community has developed various quality diversity (QD) algorithms that aim to find a balance of novel and high-performing individuals within a population. Some methods can even be considered policy generators: NEAT and HyperNEAT \cite{neat, hyperneat} use an indirect encoding (the genotype) to map to a network architecture and corresponding behavior (the phenotype).

NEAT and HyperNEAT evolve a small network into a larger network by incrementally making changes to the genotype and measuring fitness. To preserve local innovation and speciation, similar genotypes explicitly share fitness, so that an individual fitness peak can only hold so many individuals. This encourages natural selection to preserve mutations that explore novel areas of the genotype. While NEAT and HyperNEAT encourage diversity along the genotype, ADAP assumes a diverse genotype because of its uniform latent prior. Then, ADAP focuses on optimizing diversity of the behavior, or phenotype.

Focusing more on phenotype diversity, Novelty Search (NS) \cite{noveltysearch} learns individuals that have high novelty along some user defined behavioral distance metric. For example, in a maze navigation task, the behavioral characteristic could be the final resting location of the individual, and agents are selected based on how far away they end up from an archive of past individuals. Balancing novel solutions with fit ones is a problem with NS, that NS with Local Competition \cite{nslc} and MapElites \cite{mapelites} aim to solve. MapElites uses a feature descriptor of the individual (phenotypic, genotypic, or some combination), which is then bucketized for local fitness competition and speciation.

One issue with these algorithms is that they can be very sensitive to the choice of behavioral characteristic or feature descriptor \cite{qdchallenges}, and in many cases require domain knowledge for understanding which axes of diversity make sense for a given environment. On the other hand, ADAP does not require the construction of a behavioral characteristic or feature descriptor, as we instead optimize for diversity directly from the environment action space. Furthermore, QD algorithms optimize individuals via genetic algorithms or evolutionary search, while ADAP is able to directly optimize for diversity and policy credit assignment via gradient descent.

\textit{Quality Diversity Connections to Deep RL}

There are several prior works that aim to connect ideas of quality diversity with deep reinforcement learning. Like quality diversity algorithms, these methods attempt to optimize a fixed-size population or archive of policies to be distinct from each other. \cite{masood19} finds a small set of separate-weight policies that find diverse trajectories by iteratively adding one novel policy at a time to a set of existing and fully-trained policies. \cite{hong2018} uses a KL-divergence over policies; but a policy's diversity is optimized over previous SGD updates of itself, thus limiting the potential multi-modality of solutions. \cite{popdiv} optimizes for diversity of the total population via maximizing the determinant of a population distance matrix, but works best only with small populations of size three or five. Population-Based Training \cite{pbt} does not directly optimize for diversity, but instead uses different hyperparameter sets on different individuals in a population to create a training curriculum. Then, at certain intervals, the best performing individual is cloned across the population.

All of the methods in this section utilize a fixed-size population of separate-weight individuals. In contrast, our work integrates the goals of quality diversity into a memory efficient deep RL framework by simulating an entire population of agents into the weights of single generative model of policies, in which the number of unique individuals is bounded only by the capacity of the generator.

\paragraph{Option Discovery for Hierarchical RL} The option framework introduced by \cite{sutton} could be thought of as learning a generator of skills, which are temporal abstractions over actions that can be used by a downstream, higher-level controller. Recent works \cite{diayn, florensa17, valor} in option discovery learn a fixed set of skills that are discriminable by observed state or trajectory, and oftentimes learn skills without any extrinsic environmental reward. On the other hand, our method learns a continuous space of diverse and stand-alone policies trained to maximize extrinsic environmental reward and behavioral diversity -- defined as difference in actions taken given state. Additionally, requiring discriminability via state or trajectory imposes diversity constraints along a and specific task-dependent axis, and may not work in situations where state has low variability.

\paragraph{Goal-Conditioned Reinforcement Learning} Yet another way to induce diverse policy behaviors is through using goal-conditioned policies \cite{leslie, planningwithgcp, uvfa} that use a family of task-defined value or Q functions or expert trajectories \cite{gail} to incentivize diversity. These methods require structure in how to define diversity, such as defining a value function family over states \cite{uvfa}.

\paragraph{Multi-Agent Roles} Recent works generate specialize agent policies in a multi-agent setting, building on QMIX \cite{qmix}. ROMA \cite{roma} learns agent roles that are not static through agent trajectories, require optimizing several additional objectives, and are learned jointly with other roles via a joint action-value function. Similarly, MAVEN \cite{maven} optimizes the mutual information between joint agent actions and a latent variable. In comparison to these methods, ADAP has a simpler methodology, does not require generating separate agent parameters, focuses on optimizing individual rewards by learning species (not roles), and is applicable to single or multi-agent environments.

\section{Experiment 1: Can We Learn a Multi-Modal Policy Space?} \label{4}

A good generative model of policies should be able to represent a multi-modal space of behaviors. That is: different agent policies should act like they are different individuals. Our generative model uses a shared parameter set across all agents, and naively using shared parameters could result in a `jack-of-all-trades' or `average' agent -- which is precisely what we wish to avoid. In this experiment, we test whether our policy generator is able to model policies for \textit{unique individuals} in a multi-agent setting. 

\paragraph{Environment}
We test our learning $G$ in a new open-ended grid-world environment called Farmworld, that supports multi-agent interaction and partially observable observations. The idea behind Farmworld is simple: agents move about the map to gather food from various resources, such as \code{chickens} and \code{towers}. Details of the Farmworld environment are provided in the Appendix.

\paragraph{Niche Specialization Experiment}

To test the multi-modality of \textit{G}, we set up the Farmworld environment with a hidden rule specific to this experiment: when an agent spawns, it is able harness resources from either \code{towers} or \code{chickens}. However, once it gets health from one unit type, it becomes `locked-into' that unit type, and cannot gain health from other unit types. Additionally, information about an agent's `locked-into' state is not provided as part of the agent observation. We call this part of the environment the \textit{enforced specialization rule}. This rule checks if agents can behave competently within distinct \code{tower} and \code{chicken} niches. In this experiment, and all following experiments, we train agents to only optimize their own expected reward (i.e. no shared value functions or group rewards).

\paragraph{Baselines}
In addition to ADAP, we test Vanilla PPO (which is ADAP without the diversity regularization), DIAYN \cite{diayn}, and a modification of DIAYN which we will call DIAYN*. Briefly, DIAYN* uses a continuous distribution rather than a categorical distribution, and attempts to minimize the mean-squared error between the predicted `context' $\phi(s)$ and the actual context, where $\phi$ is the discriminator and $s$ is the state. We provide more details in the Appendix. For both DIAYN and DIAYN*, we augment the intrinsic reward with extrinsic environmental reward, and searched over hyperparameter settings to find the best results. In all cases, we use multiplicative and concatenation model architectures, so even the Vanilla PPO baseline could still hypothetically learn to specialize based on integrated latents.

\paragraph{Results} 

\begin{wrapfigure}{r}{0.5\linewidth}
\vspace{-0.5cm}
\includegraphics[width=1\linewidth]{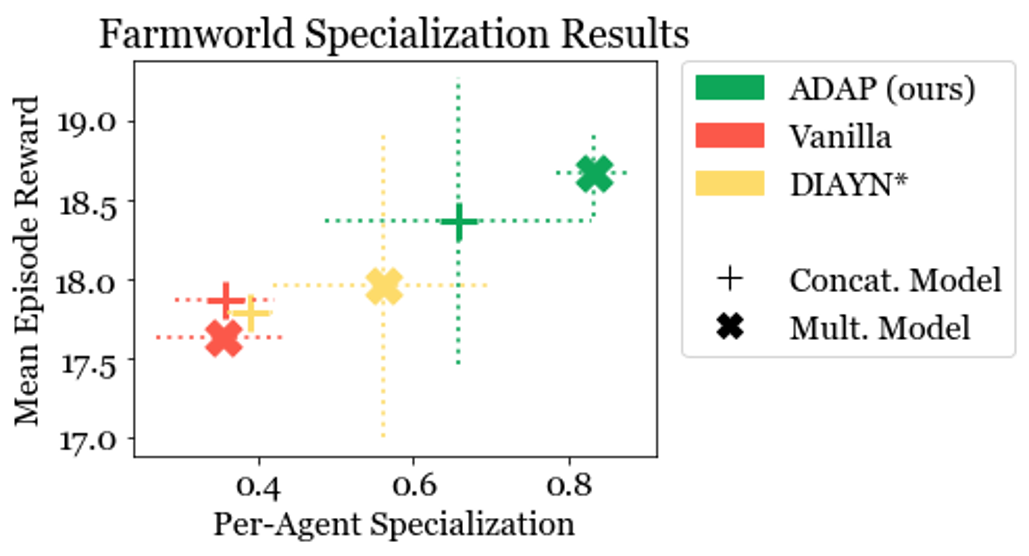}
\centering
\vspace{-0.6cm}
\caption{Reward versus Specialization}
\label{enfSpecScatter}
\vspace{-0.4cm}
\end{wrapfigure}

ADAP is able to learn a more multi-modal policy space than any of the other baselines. In other words, the policies generated by different latent vectors allowed ADAP agents to actually act like unique individuals, even though they shared the same generator parameters. Our results also indicate that using a multiplicative model can result in a higher degree of policy space multi-modality. We explain our results in more detail in the following paragraph.

We can see in Figure \ref{enfSpecScatter} that ADAP is able to attain the highest episode reward, defined by the mean reward collected over all agents. This, however, is not necessarily the most interesting point. ADAP also has the highest `per-agent specialization', which we compute as one minus the per-agent entropy of its \code{chicken} and \code{tower} attacks. That is to say, if an agent only attacks \code{towers}, it will have a specialization of one, whereas if it attacks equal numbers of \code{towers} and \code{chickens}, it will have a specialization of zero. If an agent is locked into a \code{tower} and instead attacks a \code{chicken}, this is a blunder. We qualitatively visualize the instances of blunders on the project website.

ADAP was able to \textit{encode niches} into the latent space $Z$, and found two policy sub-spaces that fell into \code{chicken} and \code{tower} niches. Indeed, during episode rollouts, ADAP agents with latents that fell into one sub-space only attempted to attack their respective unit type. On the other hand, Vanilla and DIAYN* generally failed to specialize in this way, resulting in an `average' agent that blundered often. In the cases of DIAYN* using a multiplicative model and ADAP using a concatenation model, the generators learned niches in only one of the three runs.

Note that in later experiments, we optimize the latent distribution of $G$ to fit an ablated environment, but in this experiment, we present results without latent optimization. With that being said, it could be possible to further improve the performance of ADAP and DIAYN* by optimizing the latent distribution to remove `bad apples' in the population: latents that map to less-optimal policies, or even possibly degenerate policies. Also notice that in Figure \ref{enfSpecScatter}, we have excluded the results of DIAYN to better visualize the rest of the methods, because DIAYN was an outlier with respect to the rest of the methods. The best performing DIAYN settings scored a maximum reward of 16.25 $\pm$ 0.01 and had a specialization of 0.74 $\pm$ 0.4. Possibly because intrinsic reward was not normalized across batches, DIAYN learned degenerate solutions that attempted to find states that provided information about agent contexts, essentially wedging agents against borders of the map, or between \code{chickens} and \code{towers} without actually interacting with the units themselves.

\section{Experiment 2: Adaptation to Environmental Ablations} \label{5}

In nature, differences between species and even within species lend robustness to life as a whole. It becomes less likely that any single perturbation in the environment will break the overall system. In the same manner, differences between policies can lend robustness to the policy space as a whole.

\paragraph{Experiment}

\begin{table}[ht]
\begin{tabular}{p{0.18\linewidth}| p{0.76\linewidth}}
Ablation        & Description \\
\hline
\hline
\code{Far Corner}     & 18x18 map size. Food spawns in the bottom right, agents spawn in the top left.\\
\hline
\code{Wall Barrier}    & A `rift' opens up between agents and their food. Agents must be able to navigate up and around the wall.\\
\hline
\code{Speed}          & Single agent on 2x2 map. Food health yield is set very low, so agents must be able to rapidly and consistently farm adjacent towers.\\
\hline
\code{Patience}         & Single agent on 2x2 map. Food yield is very high, but food respawn speed is low. Agents must `ration' their food so it lasts until the next respawn.\\
\hline
\code{Poison Chickens} & Agents spawn on the same map in which they were trained. However, chickens now yield \textit{negative} health. Towers still yield positive health.\\  
\hline
\code{Training Env.}     & (Not an ablation: listed for reference) 10x10 map size. Food and agents are uniformly randomly distributed.\\
\end{tabular}
\caption{Farmworld Ablations}
\label{farmAblations}
\vspace{-0.2cm}
\end{table}

We aim to test how having a diverse policy space allows us to search in latent space for policies that better fit unexpected environmental ablations. Doing so would demonstrate the robustness of a population of policies, and simultaneously provide information about different types of diversity that are learned by $G$.

To this end, we train $G$ on a normal Farmworld environment without the enforced specialization rule from Experiment 1. In this experiment, we use a Farmworld environment containing ten each of agents, \code{towers}, and \code{chickens}, randomly initialized on a 10x10 grid.

We then \textit{ablate} the environment, changing features such as map size and features, location of food sources, and even re-spawn times and food-yield. Lastly, we deploy $G$ into the ablated environment and \textit{without} changing the weights $\phi$, optimize the latent distribution for policies that are successful in the new environment, using the search algorithm described in Section \ref{2}. Ablations and descriptions are available in Table \ref{farmAblations}.

\begin{figure}[h]
\includegraphics[width=1\linewidth]{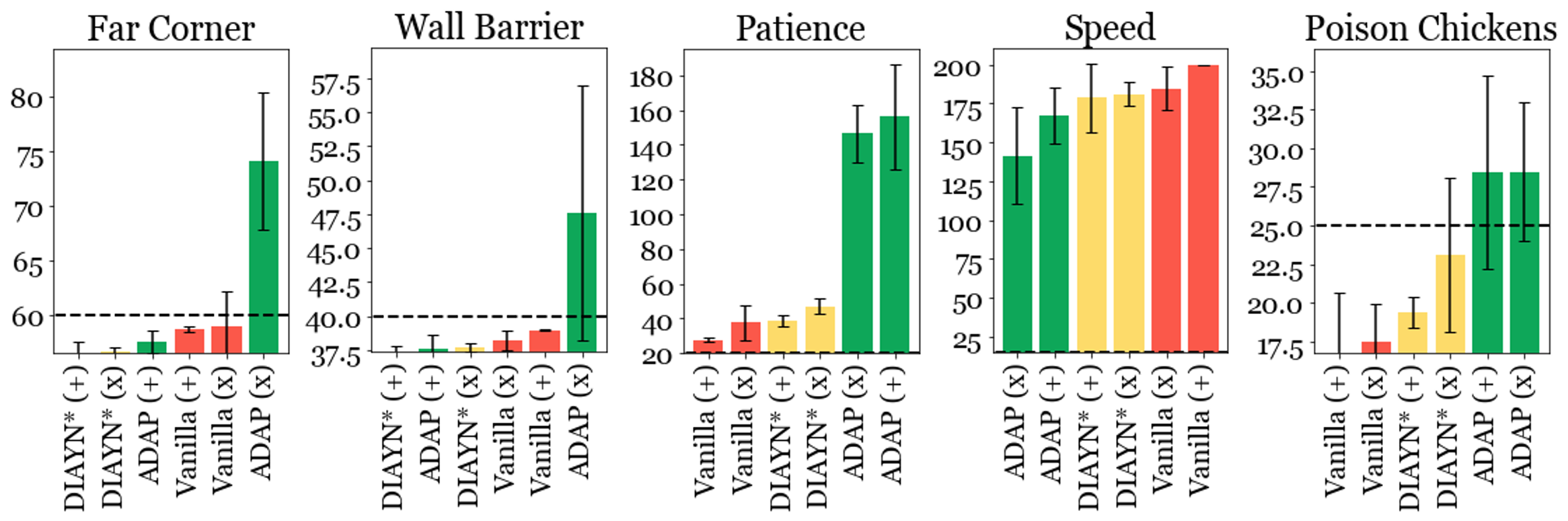}
\centering
\vspace{-0.7cm}
\caption[Lifetime After Optimization on Ablations]%
{Lifetime After Optimization on Ablations. Plots are of agent health (y-axis) after latent optimization on each ablation, and we plot the agent's initial health as a dashed black line. Error bars reflect standard deviations over three random seeds. It is possible to go below initial health, since agents can do damage to each other or eat poisonous food.}
\label{farmBars}
\vspace{-0.4cm}
\end{figure}

\begin{figure}[h]
\includegraphics[width=1\linewidth]{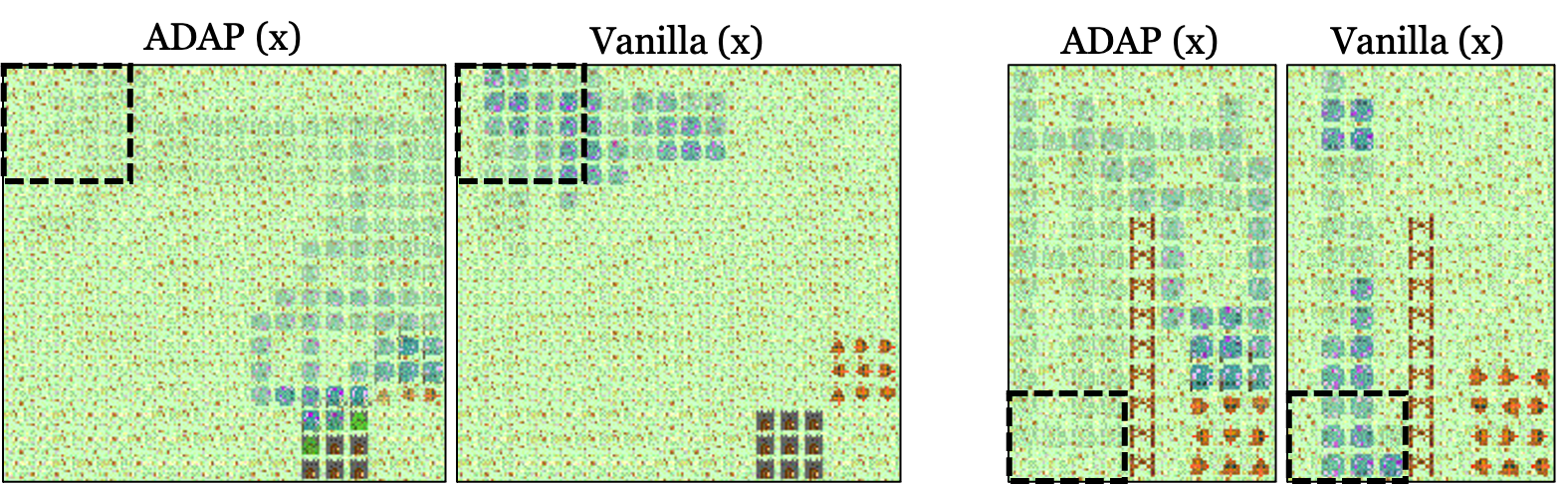}
\centering
\vspace{-0.7cm}
\caption[]%
{Adaptation on Farmworld Locomotion Ablations. Randomly chosen episode visualizations after latent optimization of \code{Far Corner} (left) and \code{Wall Barrier} (right). Chosen ADAP latents yield agent policies that navigate to the food, without otherwise modifying network parameters. Agents spawn in the dotted black outlines, and timesteps are stacked such that earlier ones are more transparent. We show 60 and 40 timesteps for \code{Far Corner} and \code{Wall Barrier} respectively.}
\label{farmAblationImg}
\vspace{-0.2cm}
\end{figure}

\begin{figure}[ht]
\includegraphics[width=1\linewidth]{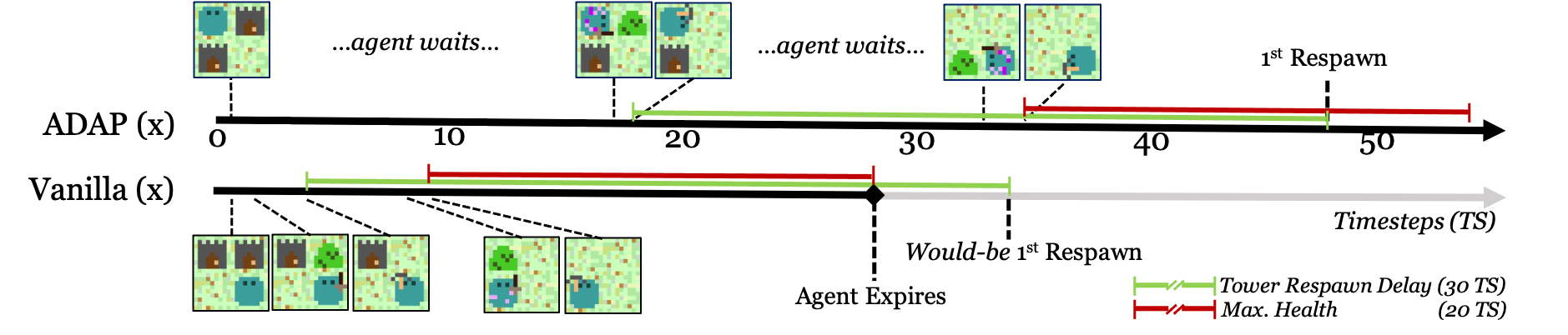}
\centering
\vspace{-0.7cm}
\caption[]%
{Adaptation on Farmworld \code{Patience} Ablation. We show episode rollouts arranged along a timeline, using agent policies after latent optimization. Notice that the Vanilla agent does not ration out the two towers. Meanwhile, the patient ADAP agent waits until its health is very low (visualized by purple and pink speckles) before getting food and is able to survive past the next tower respawn.}
\label{patientAblation}
\vspace{-0.2cm}
\end{figure}

\paragraph{Results}

Rather to our surprise, in each experiment trial, learning $G$ using ADAP created a policy space $\Pi$ containing `species' that could thrive in nearly every environmental ablation (see Figure \ref{farmBars})! In the \code{Patience} ablation (Figure \ref{patientAblation}), we see that $\Pi$ contained an agent that purposefully refrained from eating food until absolutely necessary. Perhaps equally surprising, $\Pi$ contained an agent that was able to navigate completely around the wall of \code{Wall Barrier} (Figure \ref{farmAblationImg}), without ever explicitly training on an environment in which such behavior was required. The important thing to note is that all of these behaviors are \textit{emergent from the train environment} -- a product of optimizing for both policy diversity and reward maximization while training $G$.

How is it possible that ADAP produced a policy space capable of adapting to nearly every ablation? The train environment was relatively abundant with resources scattered about a large map. Thus, there were many degrees-of-freedom in the rules of survival, and by optimizing for diversity, we found a policy space that filled these degrees-of-freedom while still yielding high reward. While these ablations reflect some of the possible axes of diversity, there are certainly more. For example, an agent's direction of `preference' does not have to be the bottom-right, as in the \code{Far Corner} ablation. Indeed, as a sanity check, we tested placing food locations in various other spots on an enlarged map, and found that for every cardinal location, there was a species of agent in $\Pi$ that could exploit that new food location. What came as a surprise was that agents also used their health indicator to diversify: since agents diversify conditional on state, species developed in which agents would prefer to go upwards when their health is high, but downwards when their health is low. This particular agent species was the one that managed to thrive in the wall barrier ablation. Similarly in the \code{Patience} ablation, ADAP learned a certain species of agent that waited until its health was low to farm a \code{tower}.

The \code{Poison Chicken} ablation was the one hold-out in which latent optimization could not find a profoundly successful species. While learning a unit preference (as in the niche specialization experiment) could be possible, it is likely that the \textit{cost} of doing so in the train environment was too great. That is to say, there would have been too large a trade-off between diversity and reward in the training environment. 

Finally, we should note that ADAP beat the Vanilla baseline in all ablations aside from \code{Speed}. We hypothesize this is because the speedy-strategy best suited the \code{Speed} ablation is also best suited for high rewards in the train environment. Since Vanilla PPO optimized for solely for expected rewards in the train environment, this is the one species that it found and perfected. DIAYN* performed similarly to Vanilla PPO, and did not learn to speciate in a manner that was successful on the majority of ablations.

\section{Experiment 3: Adaptation to Adversaries in Markov Soccer} \label{6}

\paragraph{Environment} This experiment uses Markov Soccer, introduced in \cite{markovsoccer}. Two agents, A and B, play on a gridworld and must `carry' a ball into the opposing goal to score. Agents walk in cardinal directions or \code{stand} in place. Possession is randomly initialized, and switches if one an agent bumps into the other. Actions of A and B occur on the same timestep, execution order is randomized, and each timestep ends in a draw with some $\epsilon$ probability.

Markov Soccer is an interesting environment, because the best policy for one agent depends on the policy of the other agent. As described in \cite{markovsoccer}, there exists a worse-case-optimal probabilistic policy for Markov Soccer, which maximizes the minimum possible score against any adversary. This strategy tends to be conservative, preferring to act towards a draw where a different policy could have obtained a higher score. On the other hand, non-worse-case-optimal strategies may be less conservative and may achieve very high scores against some opponents, but very low scores against others. Analogous to real soccer, different players have varying abilities and play styles, and a given player $p_1$ may be optimal against $p_2$, but not against $p_3$.

If any single policy has its drawbacks, can we instead learn an entire space of diverse policies $\Pi := \{\pi_1, \pi_2,...,\pi_{\inf}\}$, where for any opponent, we can select a policy $\pi_i \in \Pi$ that achieves the maximum score against that opponent? Ideally, this space includes the worse-case-optimal policy, as well as other more aggressive policies. Then, just as a coach might swap out a soccer player, we can mix and match our champion as suited.

\paragraph{Experiment} Can we learn a population of individuals that is holistically strong against all types of opponents? We evaluate adaptability to various adversaries using two methods. First, we test baselines and our method against a set of hand-coded soccer bots. These bots are designed to represent a wide gamut of strategies, some of which are more exploitable than others. Secondly, we evaluate each \textit{G} by playing ADAP (x), ADAP (+), Vanilla (x), and Vanilla (+) in a round-robin tournament against each other. For all evaluation methods, score is determined by \code{Wins - Losses} over the course of 1000 simulated games.

\textit{Against Hard-Coded Bots:} Each bot always starts on the left side, and the learned policy starts on the right side (although the environment is coded such that observations are side-invariant). Bot types fall into three categories: offense (bots start with possession), defense (policy starts with possession), and mixed (random starting possession). See Table \ref{botAdversaries} for more details.

\begin{table}[ht]
\begin{tabular}{p{0.16\linewidth}| p{0.09\linewidth} | p{0.65\linewidth}}
Bot & Bot Type & Bot Policy \\
\hline \hline
\code{Straight}        & Offense & Always moves \code{right}, towards the goal.\\
\hline
\code{Oscillate 0}     & Defense & Oscillates in column 0, blocking both squares adjacent the goal.\\
\hline
\code{Oscillate 1}     & Defense & Oscillates in column 1, leaving a gap in front of the goal.\\
\hline
\code{Stand}    & Defense & \code{stand}s in one square adjacent the goal, leaving the other open.\\
\hline
\code{Rule-Based}          & Mixed & Follows hand-coded heuristics.\\
\hline
\code{Random}         & Mixed & Follows a random policy.\\
\hline
\end{tabular}
\caption{Markov Soccer Bot Adversaries}
\label{botAdversaries}
\end{table}

\textit{Round-Robin Against Each Other:} Let the two generative models of policies be $G_1$ and $G_2$. Evaluation is tricky, since each generative model could map to an entire space of policies. Our goal is to evaluate the best individual policy from $G_1$ against the best individual policy from $G_2$.  To compute the score of $G_1$ against $G_2$, we select the latent vector $z_2$ for $G_2$ that maximizes the expected reward of $G_2$ against the entire family of policies of $G_1$. Then, we select the latent vector $z_1$ for $G_1$ that maximizes $G_1$’s reward against $\pi_{G_2, z_2}$. The score of $G_1$ vs $G_2$ is $\pi_{G_1, z_1}$ versus $\pi_{G_2, z_2}$.

\paragraph{Training and Baselines}

We use self-play to train both ADAP and baselines. As a baseline, we use Vanilla PPO with entropy regularization, using both concatenation and multiplicative model types. We found that using a high entropy regularization coefficient of 0.1 was necessary for training the Vanilla baseline, to prevent it collapsing into degenerate solutions. ADAP was less dependent on entropy regularization, and we used a coefficent of 0.05.

\paragraph{Results}

\begin{figure}[ht]
\includegraphics[width=1\linewidth]{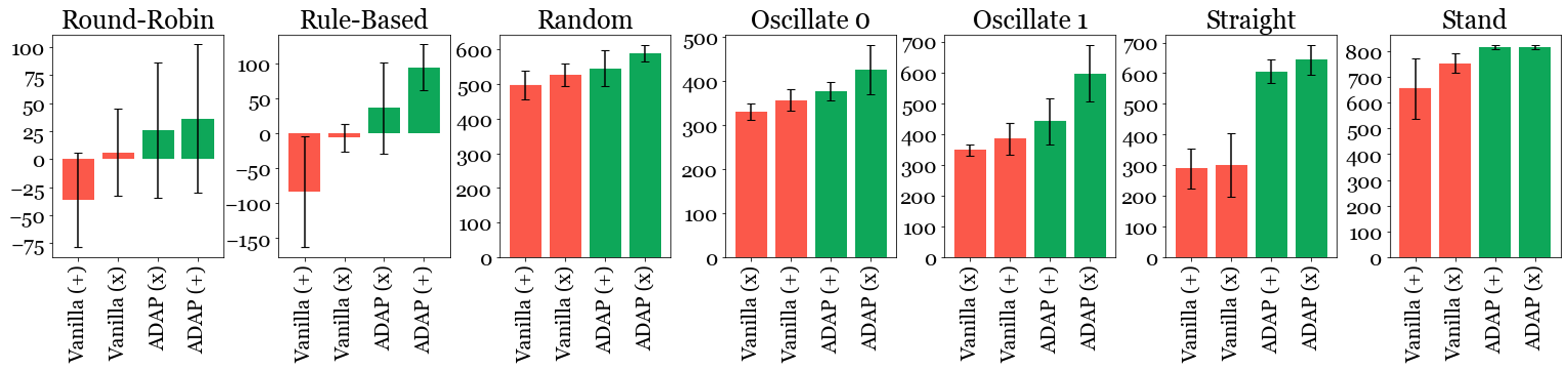}
\centering
\vspace{-0.7cm}
\caption{Results of round-robin and bot games. Score is over 1000 runs, some games prematurely end in a draw. We report standard deviation over three random seeds.}
\label{soccerBars}
\end{figure}

As in the Farmworld adaptability experiment, we see from Figure \ref{soccerBars} that ADAP is able to learn a $G$ during the \textit{train} phase that emergently contains members that are successful against a variety of unexpected adversaries - including naive bots and other policies.

Compared to Vanilla, the ADAP policy space generalizes better against the `naive' adversaries \code{Random}, \code{Oscillate 0 \& 1}, and \code{Straight}. Going back to the soccer team example, we were able to select individuals from the ADAP population that particularly exploited facets of these naive strategies. For example, against the \code{Oscillate 1} adversary, ADAP latent optimization found a member of the population that \textit{side-stepped} the oscillating adversary simply by moving to the top row, and then down to the goal. Additionally, against the \code{Straight} adversary, successful ADAP individuals stole possession by deterministically standing in-front of the opponent to block, and then moving around and into the goal. On the other hand in both of these situations, Vanilla could not find individuals that exploited the naive deterministic nature of their opponents. Instead, Vanilla individuals played as they would against themselves: cautiously and defensively, which resulted in fewer wins against the `naive' bots.

Using ADAP did not just allow us to optimize against naive opponents. ADAP learned the best $G$ in the round-robin tournament, and was the only method that was able to consistently beat our rule-based bot. It is possible that by using ADAP in training, individuals encountered a wide variety of strategies that bettered overall performance. In general, using ADAP allowed us to emergently find a policy space that exhibited strong performance against a gamut of opponents, which any single policy would not have been able to do (without knowing \textit{a priori} what type of opponent it is facing).

\section{Limitations}

\paragraph{Bad Apples} When using ADAP, not every member of the policy space is going to be an optimal policy. In fact, some generated policies might be bad apples: policies that were incentivized by the diversity regularizer to take actions that were not rewarding. Naturally, some individuals might be better or worse than others. These individuals can be removed by optimizing the latent distribution. However, the bad apples may come with a plus side. Even though they do not perform well in the current environment, they might happen to perform well in a future environment!

\paragraph{Continuous-Action Space Environments} The results presented so far focus entirely on environments with discrete categorical action spaces, in which we have observed that our diversity regularizer in Equation \ref{eq:divobjkl} empirically performs well. However, not all environments in RL use discrete action spaces - continuous action spaces are widely used in RL control tasks. While we believe that our regularizer can work in these environments, we have not rigorously tested in these environments.

\section{Conclusion}

We have presented a framework to learn a generative model of policies. Rather than learning just one policy, we aim to find as many high-performing and individually distinct policies as possible, all compressed within the parameters of our generator.

Learning a space of policies pays off in an open-ended environment such as Farmworld, in which there may be more than one path to success. We show in Section \ref{5} that we can adapt to ablations by quickly choosing `species' from our learned policy space that are successful in the new environment.

We also learn a policy space in a competitive, two-player, zero-sum game in Section \ref{6}. Here, no single deterministic policy is optimal against all adversaries. Instead, we show how to train a family of policies that can be naturally adaptable to a wide array of both challenging and naive adversaries.

Overall, we hope to show how it can be beneficial in RL settings to optimize not just for reward, but also for diversity of behavior. As environments continue to increase in complexity and open-endedness -- filled with branching paths to success -- it makes sense to learn not just one, but many, solutions.

\section{Acknowledgements}

This research was supported in part by IBM through the MIT-IBM Watson AI Lab.

\section{Appendix}

\subsection{Description of Farmworld}

Farmworld is an open-ended gridworld environment designed with two goals in mind: high customizability and support for diverse solutions. The environment is written entirely in Python, and is easily hackable. Maps can be hand-crafted, or randomly generated. At a high level, agents must traverse the environment to find resource units, and harvest health from these resources. Agents can interact directly by attacking each other, or indirectly by competing for shared and limited resources.

Units have configurable levels of health, damage, food yield, and respawn, which can be utilized to encourage learning a variety of different policies. For example, by placing low food yield on resource units (\code{chickens} and \code{towers}) and high food yield on agent units, one may incentivize direct multi-agent competition (agents must attack each other to get health). Conversely, setting a high resource unit health can encourage multi-agent co-operation - agents will have to work in parallel to mine a \code{chicken} or \code{tower} before their health runs out.

\begin{figure}[ht]
\includegraphics[width=0.8\linewidth]{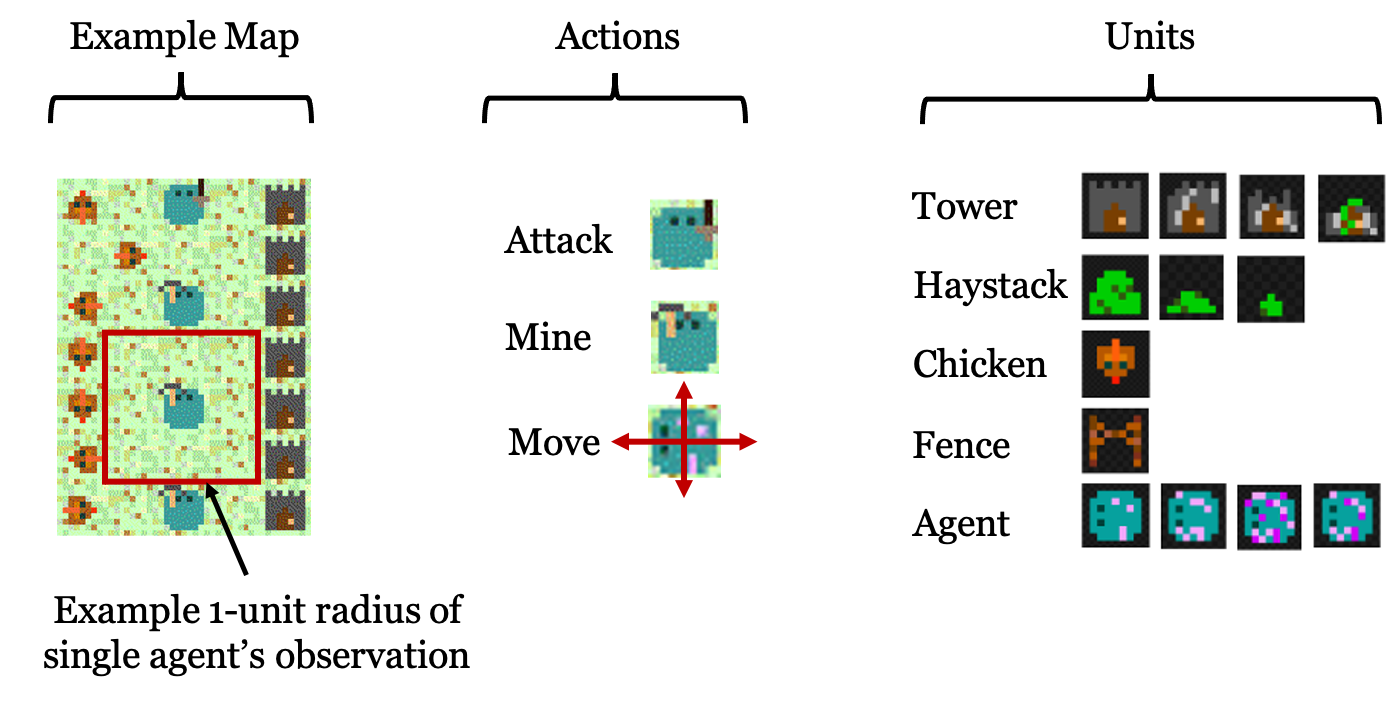}
\centering
\vspace{-0.4cm}
\caption{Illustration of Farmworld and Units}
\label{farmworldIllustration}
\vspace{-0.4cm}
\end{figure}

\paragraph{Observation Space}
Can be either RGB images, or a flattened array of unit-encoding vectors. If RGB images are used, agents `see' exactly what we see: units visibly lose health by damage patterns that appear over time, and unit orientations can be discerned by the unit `eyes' (see the Example Map in Figure \ref{farmworldIllustration}). If unit-encoding vectors are used, then all units have encoded \code{health}, \code{orientation} (by default, 0 - 3 to represent each possible cardinal direction), and unit \code{type} (e.g. 0 for ground, 1 for agents, 2 for \code{chickens}, 3 for \code{towers}, 4 for \code{fences}). Encodings are scaled to be within [0, 1].

Agents have partially-observable observations: they do not see the entire map. By default, they can see units in a L1 radius of 2 unit squares.

\paragraph{Action Space} The action space is 6-dimensional categorical, respresenting \code{up}, \code{down}, \code{right}, \code{left}, \code{attack}, \code{mine}. Actions can be added or removed as necessary.

\paragraph{Unit Pecularities} \code{Towers} and \code{Chickens} each have a corresponding non-negative \code{respawn\_time}. \code{Chickens} disappear after they get hit \code{max\_health} times by an agent \code{attack}. Unlike \code{towers}, \code{chickens} are able to move 1 square in any direction on each timestep, with \code{chicken\_move\_probability}.

\code{Tower} units are more tricky: they turn into a \code{haystack} in the same location after \code{max\_chicken\_health} of \code{attack}. \code{Haystacks} must be \code{mined} with a pickaxe, and only after \code{max\_tower\_health} will they yield food resources.

\code{Fence} units are simple: they cannot be destroyed or moved. Additionally, no units can pass through them.

\paragraph{Reward} Agents get an individual reward of 0.1 for each timestep that they are alive. If an agent's health reaches 0, it is removed from the map and other agents can carry on as normal. The entire episode ends when \code{max\_episode\_timesteps} is reached, or when no more agents are alive on the map.


\subsection{Algorithm Pseudocode}

\begin{algorithm}
\caption{ADAP with PPO}\label{ppo_div}
\begin{algorithmic}[1]
    \State{$m$ the number of sampled latents in diversity estimation}
    \State{$n$ the number of sampled states in diversity estimation}
    \State{$k$ latent vector size}
    \State{$\alpha$ diversity regularization coefficient}

    \For{\text{for iteration = 1, 2, ...}}
        \State Let $B$ be an empty batch of (s, a, r) tuples
        \For{\text{actor $a$ = 1, 2, ..., $N$}}
            \State \text{Sample latent $z$ from latent distribution}
            \State \text{$B \leftarrow$ Run policy $\pi(\cdot|\theta_{old};z)$ in environment for $T$ steps}
            \State \text{Compute advantage estimates $\hat{A_1},...,\hat{A_T}$}
        \EndFor
        \State Sample $M \in \mathbb{R}^{m \times k}$ from the latent distribution \Comment \text{latent matrix}
        \State Sample a batch $S$ of $n$ states from $B$
        \State $L_{div} \leftarrow 0$
        \For {\text{$i = 1, 2, ..., m - 1$}}
            \For {\text{$j = i + 1, i + 2, ..., m$}}
                \State $L_{div} \leftarrow L_{div} + \frac{1}{n} \sum_{s \in S} D_{KL}(\pi(s|\theta_{old}, M^{(i)}), \pi(s|\theta_{old}, M^{(j)}))$
            \EndFor
        \EndFor
        \State $L_{div} \leftarrow \frac{2}{m(m-1)} L_{div}$  \Comment \text{Scale by number of policy-distance pairs}
        \State \text{Maximize $L_{PPO} - \alpha L_{div}$ w.r.t. $\theta$} via SGD.
        \State $\theta_{old} \leftarrow \theta$
    \EndFor
\end{algorithmic}
\end{algorithm}

\begin{algorithm}
\caption{Latent Distribution Optimization}\label{lso}
\begin{algorithmic}[1]
    \State \textbf{Input:} $g$ the number of optimization generations
    \State \textbf{Input:} $E$ an environment
    \State \textbf{Input:} $G$ a policy generator
    \State \textbf{Input:} $Z$ a latent distribution with dimension $k$

    \State \textbf{Initialize:} \texttt{best} $\leftarrow$ descending sorted array
    \For{$i = 1, 2, ..., g$}
        \State $\code{explor} \sim$ Unif([0, 1])
        \State $r \sim$ Unif([0, 1])
        \If{$(\code{explor} \leq 0.5$ \textbf{and} $i \leq \frac{3}{4}g$)  \textbf{or} len(\code{best}) $\leq 10$}
            \If{$r \leq 0.5$ \textbf{or} len(\code{best}) $\leq 10$ }
                \State $z \sim Z$ \Comment \text{Random Sampling}
            \Else
                \State $z \leftarrow$ sample(\code{best[0:10]})
                \State $z \leftarrow$ $z$ + project$_{Z}(\text{Unif[-0.1, 0.1]}^k)$ \Comment \text{Mutation}
            \EndIf
        \Else
            \If{$r \leq 0.5$}
                \State $z \leftarrow$ sample(\code{best[0:10]}) \Comment \text{Replication}
            \Else
                \State $z \leftarrow$ pop(\code{best[0:10]}) \Comment \text{Pruning}
            \EndIf
        \EndIf

        \State \textit{score} $\leftarrow$ Reward from running $\pi_{G, z}$ on $E$
        \State \code{best.push(\text{$z$)}} with key \textit{score}
    \EndFor
    \State \textbf{Return} \code{best[0]}
\end{algorithmic}
\end{algorithm}

\newpage

\subsection{Baselines}

DIAYN \cite{diayn} originally attempts to maximize the mutual information between the state and a discrete categorical latent vector by optimizing an intrinsic reward generated from discriminator error. We wanted to make a comparison of DIAYN to ADAP in which both methods used continuous latents to find a potentially unbounded number of niches. To this end, we augmented DIAYN and called this DIAYN*. In DIAYN*, we train the discriminator to regress the latent, rather than predict the latent category. We add this intrinsic reward to the extrinsic environmental reward, giving us the new reward function $r'$:
$$r'_t = err_t + r_t$$
where
$$err_t = -\alpha (q_{\phi}(s_t) - z)^2 - \text{mean}(err_{batch})$$
$z$ is the latent vector, $\text{mean}(err_{batch})$ is the mean discriminator error across the update batch, and $\alpha$ is the scaling of the intrinsic reward (generally set at 0.05). We subtract by the batch mean so that on average, the expected agent reward equals only what is provided by the extrinsic environment. Otherwise, original DIAYN and DIAYN* stuggled with balancing dense extrinsic environmental rewards from the experiment with the intrinsic discriminator reward. In our niche specialization experiment, we also experimented with the canonical DIAYN. In this implementation, we use categorical contexts and we add extrinsic reward directly to intrinsic discriminator reward. As mentioned in the paper, this method did not perform well in our Farmworld Niche Specialization experiment.

Finally, we treat DIAYN and DIAYN* like a generative model of policies (since we are not trying to learn options). To do so, we keep $z$ fixed throughout an agent episode. Included in the website are toy experiments that benchmark our implementations of DIAYN*.

\newpage

\subsection{Smoothing Parameter $b$}
In continuous domains with action distribution $\mathcal{N}(\mu, \sigma)$, we observed that the KL-divergence in Equation \ref{eq:divobjkl} may encourage very low $\sigma$ values early in ADAP training. To solve this, we used standard deviation $\sigma' = \sigma + b$, where $b$ is a small constant (ex: 0.05). We similarly use the smoothing parameter $b$ in the discrete action spaces, but have not tested whether or not it is necessary in these situations.

\subsection{Training Hyperparameters}

Unless otherwise mentioned, we used optimized our policies using a clipped PPO surrogate objective with learning rate 3e-4. Advantages were computed using Generalized Advantage Estimation, with a $\gamma$ discount factor of 0.99, a $\lambda$ smoothing parameter of 1, and a gradient clip of 0.5. We use the RLLib \cite{rllib} framework for training, using their default PPO configuration. For all experiments, we use concatenation and multiplicative model architectures as seen in the main paper. Importantly, we always use \textit{separate value and policy networks}. Attempts to combine these networks generally resulted in non-diverse policy spaces, which we believe is a result of the importance of the value function in recognizing the differing expected rewards conditional on each latent from the latent space.

For multi-agent environments, batch sizes are always in \textit{agent} steps, rather than in \textit{environment} steps. Thus, if there are 40 agents in an environment, then 1 environment step is 40 agent steps.

To optimize our diversity regularization objective, we use parameters $m = 10, b = 30, k = 3$, as detailed in \ref{ppo_div}. However, preliminary investigation into the effect of these hyperparameters indicates that it is possible to get away with even smaller samples of latent vectors and states, while still effectively optimizing for a diverse policy manifold.

Unless otherwise specified we use a diversity regularizer coefficient on our novel objective of coefficient of 0.1 in CartPole, 0.2 in Farmworld, 0.2 in Markov Soccer, and 0.5 in MultiGoal. When using DIAYN and DIAYN*, we found that a small intrinsic reward coefficient of 0.05 was best. Anything beyond that, and DIAYN and DIAYN* had issues optimizing for actual extrinsic reward.

For all methods, we generally use an 0.05 entropy coefficient, except in Markov Soccer in which we also run Vanilla PPO with 0.1 entropy coefficient and were able to achieve slightly stronger performance. In our Markov Soccer experiment, we report the average of these two Vanilla results.

\newpage

\begin{table}
\centering
\begin{tabular}{ll}
Batch size               & 4000         \\
Minibatch size           & 400         \\
SGD iterations per batch & 10           \\
Training epochs          & 200  \\
Hidden dimension         & 16           \\
Value Activations        & ReLU         \\
Policy Activations       & Tanh         \\
\end{tabular}
\caption{CartPole}
\end{table}

\begin{table}
\centering
\begin{tabular}{ll}
Batch size               & 4000         \\
Minibatch size           & 400         \\
SGD iterations per batch & 10           \\
Training epochs          & 500  \\
Hidden dimension         & 32           \\
Value Activations        & Tanh (ReLU for DIAYN*)        \\
Policy Activations       & Tanh         \\
\end{tabular}
\caption{Multi-Agent MultiGoal}
\end{table}

\begin{table}
\centering
\begin{tabular}{ll}
Batch size               & 8000         \\
Minibatch size           & 8000         \\
SGD iterations per batch & 10           \\
Training epochs          & 10 thousand  \\
Hidden dimension         & 64           \\
Value Activations        & Tanh  (ReLU for DIAYN*)        \\
Policy Activations       & Tanh         \\
\end{tabular}
\caption{Niche Specialization and Farmworld Ablation Experiment}
\end{table}

\begin{table}
\centering
\begin{tabular}{ll}
Batch size               & 8000         \\
Minibatch size           & 8000         \\
SGD iterations per batch & 10           \\
Training epochs          & 10 thousand  \\
Hidden dimension         & 64           \\
Value Activations        & Tanh         \\
Policy Activations       & Tanh         \\
GAE lambda               & 0.95         \\
GAE gamma                & 0.9         
\end{tabular}
\caption{Markov Soccer}
\end{table}

\bibliography{main}

\begin{thebibliography}{10}

\bibitem{sutton}
Richard~S. Sutton, Doina Precup, and Satinder Singh.
\newblock Between mdps and semi-mdps: A framework for temporal abstraction in
  reinforcement learning.
\newblock {\em Artificial Intelligence}, 112(1):181--211, 1999.

\bibitem{diayn}
Benjamin {Eysenbach}, Abhishek {Gupta}, Julian {Ibarz}, and Sergey {Levine}.
\newblock {Diversity is All You Need: Learning Skills without a Reward
  Function}.
\newblock {\em arXiv e-prints}, page arXiv:1802.06070, February 2018.

\bibitem{florensa17}
Carlos {Florensa}, Yan {Duan}, and Pieter {Abbeel}.
\newblock {Stochastic Neural Networks for Hierarchical Reinforcement Learning}.
\newblock {\em arXiv e-prints}, page arXiv:1704.03012, April 2017.

\bibitem{qualitydiv}
Justin~K. Pugh, Lisa~B. Soros, and Kenneth~O. Stanley.
\newblock Quality diversity: A new frontier for evolutionary computation.
\newblock {\em Frontiers in Robotics and AI}, 3:40, 2016.

\bibitem{markovsoccer}
Michael~L. Littman.
\newblock Markov games as a framework for multi-agent reinforcement learning.
\newblock In {\em Proceedings of the Eleventh International Conference on
  International Conference on Machine Learning}, ICML'94, page 157–163, San
  Francisco, CA, USA, 1994. Morgan Kaufmann Publishers Inc.

\bibitem{openaigym}
Greg {Brockman}, Vicki {Cheung}, Ludwig {Pettersson}, Jonas {Schneider}, John
  {Schulman}, Jie {Tang}, and Wojciech {Zaremba}.
\newblock {OpenAI Gym}.
\newblock {\em arXiv e-prints}, page arXiv:1606.01540, June 2016.

\bibitem{hypernets}
David {Ha}, Andrew {Dai}, and Quoc~V. {Le}.
\newblock {HyperNetworks}.
\newblock {\em arXiv e-prints}, page arXiv:1609.09106, September 2016.

\bibitem{ppo}
John {Schulman}, Filip {Wolski}, Prafulla {Dhariwal}, Alec {Radford}, and Oleg
  {Klimov}.
\newblock {Proximal Policy Optimization Algorithms}.
\newblock {\em arXiv e-prints}, page arXiv:1707.06347, July 2017.

\bibitem{rllib}
Eric Liang, Richard Liaw, Robert Nishihara, Philipp Moritz, Roy Fox, Ken
  Goldberg, Joseph Gonzalez, Michael Jordan, and Ion Stoica.
\newblock {RL}lib: Abstractions for distributed reinforcement learning.
\newblock In Jennifer Dy and Andreas Krause, editors, {\em Proceedings of the
  35th International Conference on Machine Learning}, volume~80 of {\em
  Proceedings of Machine Learning Research}, pages 3053--3062. PMLR, 10--15 Jul
  2018.

\bibitem{neat}
Kenneth~O. Stanley and Risto Miikkulainen.
\newblock Evolving neural networks through augmenting topologies.
\newblock {\em Evolutionary Computation}, 10(2):99--127, 2002.

\bibitem{hyperneat}
Kenneth~O. Stanley, David~B. D'Ambrosio, and Jason Gauci.
\newblock {A Hypercube-Based Encoding for Evolving Large-Scale Neural
  Networks}.
\newblock {\em Artificial Life}, 15(2):185--212, 04 2009.

\bibitem{noveltysearch}
Joel Lehman and Kenneth~O. Stanley.
\newblock Abandoning objectives: Evolution through the search for novelty
  alone.
\newblock {\em Evolutionary Computation}, 19(2):189--223, 2011.

\bibitem{nslc}
Joel Lehman and Kenneth Stanley.
\newblock Evolving a diversity of creatures through novelty search and local
  competition.
\newblock pages 211--218, 01 2011.

\bibitem{mapelites}
Jean-Baptiste {Mouret} and Jeff {Clune}.
\newblock {Illuminating search spaces by mapping elites}.
\newblock {\em arXiv e-prints}, page arXiv:1504.04909, April 2015.

\bibitem{qdchallenges}
Justin~K. Pugh, L.~B. Soros, Paul~A. Szerlip, and Kenneth~O. Stanley.
\newblock Confronting the challenge of quality diversity.
\newblock In {\em Proceedings of the 2015 Annual Conference on Genetic and
  Evolutionary Computation}, GECCO '15, page 967–974, New York, NY, USA,
  2015. Association for Computing Machinery.

\bibitem{masood19}
Muhammad~A. {Masood} and Finale {Doshi-Velez}.
\newblock {Diversity-Inducing Policy Gradient: Using Maximum Mean Discrepancy
  to Find a Set of Diverse Policies}.
\newblock {\em arXiv e-prints}, page arXiv:1906.00088, May 2019.

\bibitem{hong2018}
Zhang-Wei {Hong}, Tzu-Yun {Shann}, Shih-Yang {Su}, Yi-Hsiang {Chang}, and
  Chun-Yi {Lee}.
\newblock {Diversity-Driven Exploration Strategy for Deep Reinforcement
  Learning}.
\newblock {\em arXiv e-prints}, page arXiv:1802.04564, February 2018.

\bibitem{popdiv}
Jack {Parker-Holder}, Aldo {Pacchiano}, Krzysztof {Choromanski}, and Stephen
  {Roberts}.
\newblock {Effective Diversity in Population Based Reinforcement Learning}.
\newblock {\em arXiv e-prints}, page arXiv:2002.00632, February 2020.

\bibitem{pbt}
Max {Jaderberg}, Valentin {Dalibard}, Simon {Osindero}, Wojciech~M.
  {Czarnecki}, Jeff {Donahue}, Ali {Razavi}, Oriol {Vinyals}, Tim {Green}, Iain
  {Dunning}, Karen {Simonyan}, Chrisantha {Fernando}, and Koray {Kavukcuoglu}.
\newblock {Population Based Training of Neural Networks}.
\newblock {\em arXiv e-prints}, page arXiv:1711.09846, November 2017.

\bibitem{valor}
Joshua {Achiam}, Harrison {Edwards}, Dario {Amodei}, and Pieter {Abbeel}.
\newblock {Variational Option Discovery Algorithms}.
\newblock {\em arXiv e-prints}, page arXiv:1807.10299, July 2018.

\bibitem{leslie}
Leslie~Pack Kaelbling.
\newblock Learning to achieve goals.
\newblock In {\em IN PROC. OF IJCAI-93}, pages 1094--1098. Morgan Kaufmann,
  1993.

\bibitem{planningwithgcp}
Soroush {Nasiriany}, Vitchyr~H. {Pong}, Steven {Lin}, and Sergey {Levine}.
\newblock {Planning with Goal-Conditioned Policies}.
\newblock {\em arXiv e-prints}, page arXiv:1911.08453, November 2019.

\bibitem{uvfa}
Tom Schaul, Daniel Horgan, Karol Gregor, and David Silver.
\newblock Universal value function approximators.
\newblock In Francis Bach and David Blei, editors, {\em Proceedings of the 32nd
  International Conference on Machine Learning}, volume~37 of {\em Proceedings
  of Machine Learning Research}, pages 1312--1320, Lille, France, 07--09 Jul
  2015. PMLR.

\bibitem{gail}
Yiming {Ding}, Carlos {Florensa}, Mariano {Phielipp}, and Pieter {Abbeel}.
\newblock {Goal-conditioned Imitation Learning}.
\newblock {\em arXiv e-prints}, page arXiv:1906.05838, June 2019.

\bibitem{qmix}
Tabish {Rashid}, Mikayel {Samvelyan}, Christian {Schroeder de Witt}, Gregory
  {Farquhar}, Jakob {Foerster}, and Shimon {Whiteson}.
\newblock {QMIX: Monotonic Value Function Factorisation for Deep Multi-Agent
  Reinforcement Learning}.
\newblock {\em arXiv e-prints}, page arXiv:1803.11485, March 2018.

\bibitem{roma}
Tonghan {Wang}, Heng {Dong}, Victor {Lesser}, and Chongjie {Zhang}.
\newblock {ROMA: Multi-Agent Reinforcement Learning with Emergent Roles}.
\newblock {\em arXiv e-prints}, page arXiv:2003.08039, March 2020.

\bibitem{maven}
Anuj {Mahajan}, Tabish {Rashid}, Mikayel {Samvelyan}, and Shimon {Whiteson}.
\newblock {MAVEN: Multi-Agent Variational Exploration}.
\newblock {\em arXiv e-prints}, page arXiv:1910.07483, October 2019.

\end{thebibliography}

\appendix

\end{document}